\documentclass[letterpaper]{article} 
\usepackage{aaai2026}  
\usepackage{times}  
\usepackage{helvet}  
\usepackage{courier}  
\usepackage[hyphens]{url}  
\usepackage{graphicx} 
\usepackage{natbib}  
\usepackage{caption} 
\usepackage{algorithm}
\usepackage{algorithmic}
\usepackage{amsmath}
\usepackage{booktabs} 
\usepackage{multirow, bm}

\usepackage{newfloat}
\usepackage{listings}
\DeclareCaptionStyle{ruled}{labelfont=normalfont,labelsep=colon,strut=off} 
\floatstyle{ruled}
\newfloat{listing}{tb}{lst}{}
\floatname{listing}{Listing}
\title{Accelerating Controllable Generation via Hybrid-grained Cache}
\author{
    Lin Liu\textsuperscript{\rm 1},
    Huixia Ben\textsuperscript{\rm 2}\thanks{Corresponding author.},
    Shuo Wang\textsuperscript{\rm 1},
    Jinda Lu\textsuperscript{\rm 1},
    Junxiang Qiu\textsuperscript{\rm 1},
    Shengeng Tang\textsuperscript{\rm 3},
    Yanbin Hao\textsuperscript{\rm 3}
}

\affiliations{
    \{liulin0725,lujd,qiujx\}@mail.ustc.edu.cn, huixiaben@mail.hfut.edu.cn\\
    shuowang.edu@gmail.com, tsg1995@mail.hfut.edu.cn, haoyanbin@hotmail.com\\
    \textsuperscript{\rm 1}University of Science and Technology of China\\
    \textsuperscript{\rm 2}Anhui University of Science and Technology\\
    \textsuperscript{\rm 3}Hefei University of Technology

}

\usepackage{bibentry}

\begin{document}

\maketitle

\begin{abstract}
Controllable generative models have been widely used to improve the realism of synthetic visual content. However, such models must handle control conditions and content generation computational requirements, resulting in generally low generation efficiency. To address this issue, we propose a \textbf{H}ybrid-\textbf{G}rained \textbf{C}ache (\textbf{HGC}) approach that reduces computational overhead by adopting cache strategies with different granularities at different computational stages. Specifically, (1) we use a coarse-grained cache (block-level) based on feature reuse to dynamically bypass redundant computations in encoder-decoder blocks between each step of model reasoning. (2) We design a fine-grained cache (prompt-level) that acts within a module, where the fine-grained cache reuses cross-attention maps within consecutive reasoning steps and extends them to the corresponding module computations of adjacent steps. These caches of different granularities can be seamlessly integrated into each computational link of the controllable generation process. We verify the effectiveness of HGC on four benchmark datasets, especially its advantages in balancing generation efficiency and visual quality. For example, on the COCO-Stuff segmentation benchmark, our \textbf{HGC} significantly reduces the computational cost (MACs) by 63\% (from 18.22T $\to$ \textbf{6.70}T↓), while keeping the loss of semantic fidelity (quantized performance degradation) within 1.5\%.
\end{abstract}


\section{Introduction}
\begin{figure}[t]
  \centering
   \includegraphics[width=1\linewidth]{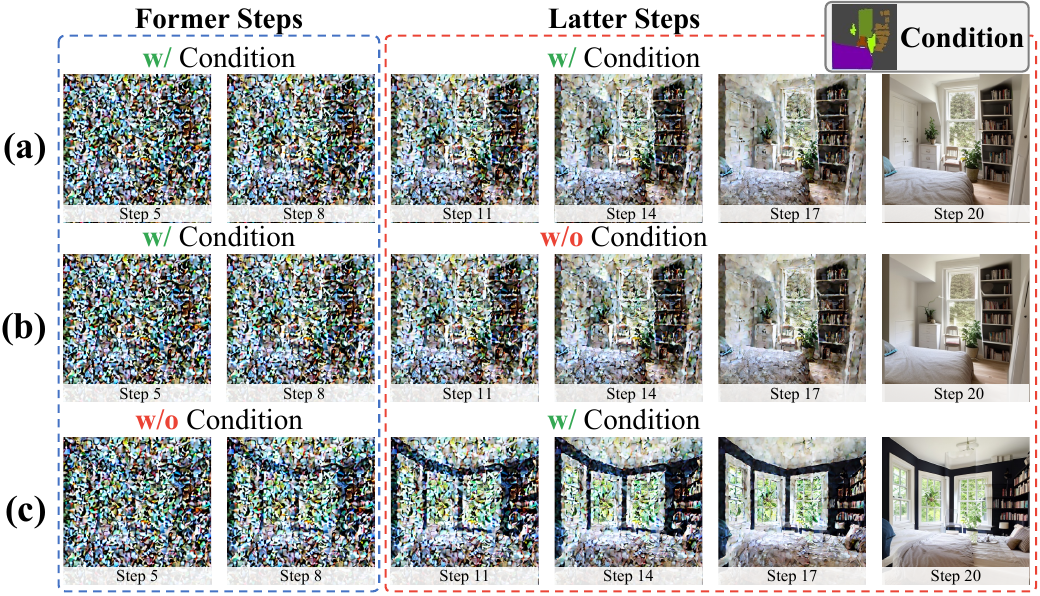}
   \caption{We visualize the intermediate results of adding conditions at different steps in the controllable generative model: (a) adding throughout steps, (b) adding only in the first ten steps, and (c) adding only after ten steps.
   }
   \label{fig:blockview}
\end{figure}

Controllable generative models \cite{zhang2023adding,li2024controlnet++} incorporate conditional information into the content generation process, enabling more precise and tailored outputs compared to traditional generative methods. Thus, these models have gained widespread adoption \cite{tang2025gloss,tang2025sign,zhang2025enhancing}. However, during the denoising generation process, these models must simultaneously compute two critical components: the control module and the generative module. This dual computation significantly increases computational complexity and inference latency, leading to slower generation speeds and reduced usability.

To investigate the impact of various components in the controllable generative models on generation efficiency and quality, we visualize the intermediate results of incorporating conditions at different stages of the model, as shown in Figure \ref{fig:blockview}. Where the model (a) applies conditions throughout all steps, (b) applies conditions only in the first ten steps, and (c) applies conditions only after the first ten steps.

Comparing (a) and (b), the image structure increasingly aligns with the conditional image during the latter stages of the model's denoising process, indicating lower computational demands at this point. However, comparing (a) and (c) reveals that omitting the conditional image constraint early on results in significant differences in the final output. Additionally, the discrepancies between (b) and (c) underscore the critical influence of when the conditional image is introduced. This suggests that establishing the semantic structure early is essential, though its importance diminishes significantly during the later stages of detail generation. These results show that there are a lot of redundant calculations in controllable generative models, and not every step requires the introduction of conditional calculations.

To reduce the computational complexity of the model, recent methods have introduced caching mechanisms. For example, Faster Diffusion \cite{li2023faster} employs temporal-aware optimization, leveraging the gradual evolution of encoder features compared to rapidly changing decoder features across timesteps. It selectively skips encoder computations and reuses cached decoder outputs. DeepCache \cite{ma2024deepcache} targets structural redundancy in denoising paths, exploiting deep-layer feature similarity to bypass recomputation through intelligent caching while preserving shallow-layer updates. T-GATE \cite{zhang2024cross} accelerates attention mechanisms by caching stabilized cross-attention outputs after convergence, minimizing redundant calculations. However, directly implementing caching in this model may compromise control fidelity, as the control module interacts closely with the generative module, requiring intensive computations. Research \cite{karras2022elucidating} indicates that the generative module’s computational demands surge in later stages, focusing on refining fine-grained details during the final denoising steps. To address this, we propose \textbf{H}ybrid-\textbf{g}rained \textbf{C}aches (\textbf{HGC}), which combines block-level caching for stage-wide efficiency with prompt-level caching for granular optimization.

First, we introduce a \textbf{block-level caching} strategy to speed up the generation process of the controllable generative model. Specifically, we compute the similarity between block outputs to pinpoint the moment when the structure changes the most, and cache the state at this critical moment to preserve the intermediate representation of the former stage (where the conditional image has the greatest impact). Reusing these cached blocks in subsequent steps can eliminate redundant computations in stages that focus on detail refinement rather than structure redefinition. It reduces computational overhead by avoiding repeated processing of the conditional image in all stages, while ensuring that the final output is consistent with the expected guidance by maintaining early semantic integrity. These operations are used in both the control module and the generative module, but in the generative module, we introduce two control ratios to adjust the cache strength between different modules and denoising stages.
Although block-level caching reduces redundant computation by reusing early structural representations, it still requires processing the entire block computation (including the cross-attention layer and multi-layer perceptron (MLP)). Therefore, to further reduce the computational cost, we propose a \textbf{prompt-level caching} strategy that achieves intra-block performance savings by focusing on the cross-attention mechanism. By analyzing the similarities between cross-attention outputs of adjacent blocks, we identify moments of minimal change and cache these states at key points, enabling the network to skip repeated computations without sacrificing image quality. This cue-level cache complements the block-level cache by storing and reusing cross-attention states, reducing the need to recompute these outputs and their associated multi-layer perceptrons (MLPs), and further reducing computational overhead while maintaining early semantic fidelity.

We define block-level caching as a coarse-grained cache, and correspondingly, prompt-level caching is defined as a fine-grained cache. These caches, with varying granularities, seamlessly integrate into each stage of the controllable generation process to enhance computational efficiency. Our contributions are threefold:
\begin{itemize}
\item We propose a coarse-grained cache that adjusts intervals based on feature similarity across steps, reducing overhead in structural synthesis while keeping quality.
\item We design a fine-grained cache that reuses stable cross-attention maps across denoising steps using temporal similarity to skip redundant steps and keep precision.
\item Our \textbf{HGC} reduces MACs by approximately 63\% across diverse datasets and tasks, achieving comparable quality while significantly boosting model speed.
\end{itemize}

\section{Related Work}
In this section, we briefly introduce the controllable generative models and the related acceleration methods.

\subsection{Controllable Generative Models}

Ho et al. \cite{ho2020denoising} introduced the Denoising Diffusion Probabilistic Model (DDPM), laying a robust foundation for iterative denoising in diffusion-based generative models. Dhariwal and Nichol \cite{dhariwal2021diffusion} added classifier guidance to direct sampling toward specific attributes, while Ho et al. \cite{ho2022classifier} advanced this with classifier-free guidance, embedding conditions directly into the model for enhanced flexibility and sample quality. For spatially conditioned synthesis, Zhang et al. \cite{zhang2023adding} extended text-to-image diffusion models with ControlNet, enabling precise geometric control, later improved by ControlNet++ \cite{li2024controlnet++} through consistency feedback to minimize artifacts. In text-to-image generation, latent diffusion models like Stable Diffusion \cite{rombach2022high} and Imagen \cite{saharia2022photorealistic} achieved high-fidelity synthesis by embedding text prompts in compact latent spaces, balancing semantic control and efficiency. Video Diffusion Models (VDM) \cite{ho2022video} adapted these methods for temporally consistent video synthesis, supporting text or frame-based conditioning for dynamic storytelling. In 3D generation, Poole et al. \cite{poole2022dreamfusion} proposed Score Distillation Sampling (SDS), using 2D diffusion priors to optimize 3D assets like neural radiance fields. Domain-specific applications, such as MedSegDiff \cite{wu2024medsegdiff}, customized diffusion for medical image segmentation with anatomical constraints.

\begin{figure*}[t]
  \centering
   \includegraphics[width=1\linewidth]{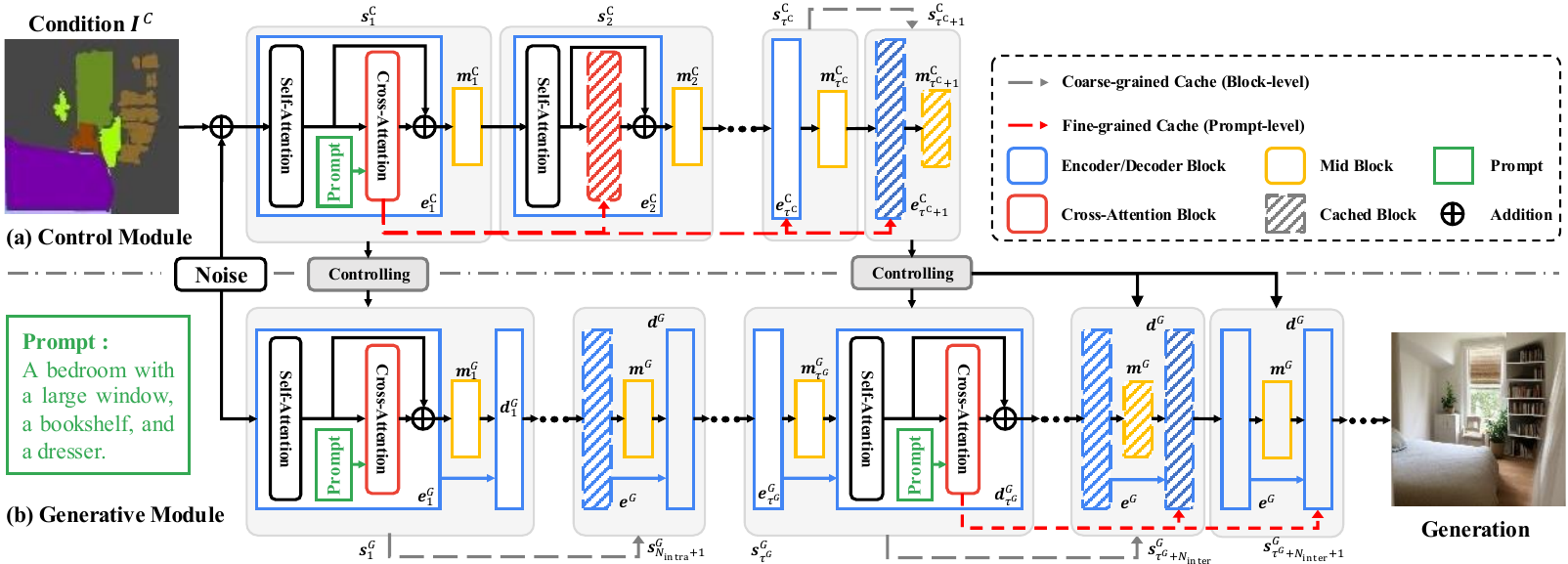}
   \caption{Controllable Generation with our \textbf{H}ybrid-\textbf{g}rained \textbf{C}aches (\textbf{HGC}),
   where the coarse-grained cache performs either full or partial caching of blocks across different steps. The fine-grained cache is governed by the gate step, which activates the cache of cross-attention maps at their respective steps.
   }
   \label{fig:pipeline}
\end{figure*}

\subsection{Model Acceleration}

Common methods for accelerating in diffusion models are grouped into four categories: \textbf{pruning}, \textbf{quantization}, \textbf{efficient sampling}, and \textbf{caching}.
\textbf{Pruning} simplifies models by removing less essential components while maintaining performance.
It is divided into unstructured pruning \cite{dong2017learning,zhu2025enhancing}, which masks individual parameters, and structured pruning \cite{liu2021group,wang2020large}, which removes larger structures such as layers or filters. 
For example, DiP-GO \cite{zhu2025dip} uses a plugin pruner to optimize constraints for better synthesis quality, and DaTo \cite{zhang2024token} dynamically prunes low-variance tokens to enhance temporal feature dynamics in self-attention.
\textbf{Quantization} compresses models by representing weights and activations in lower-bit formats. Key approaches include Quantization-Aware Training \cite{bhalgat2020lsq,DBLP:conf/nips/ZhuZ00HZ24}, which integrates quantization during training, and Post-Training Quantization (PTQ) \cite{li2021brecq,DBLP:journals/corr/abs-2507-03657}, which quantizes pre-trained models without retraining. For example, Q-Diffusion \cite{li2023qdiffusion} improves calibration with time step-aware sampling and a noise-prediction quantizer.
\textbf{Efficient sampling} 
encompasses two paradigms: retraining-based optimization and sampling-algorithm enhancement. Retraining methods, such as knowledge distillation \cite{salimans2022progressive,DBLP:conf/mm/ZhuZTWHZ24,wang2018connectionist}, modify architectures for fewer-step generation but require additional training resources. Training-free methods refine sampling dynamics. For instance, DDIM \cite{song2020denoising} accelerates inference with non-Markovian deterministic trajectories, while DPM-Solver \cite{lu2022dpm} uses high-order solvers to reduce steps theoretically. Consistency models \cite{song2023consistency,DBLP:conf/aaai/ZhuWLHL024,guo2019dense} enable single-step sampling via learned transition mappings.
\textbf{Caching} accelerates generation by reusing computations. It falls into two categories: rule-based methods \cite{selvaraju2024fora,ma2024deepcache,qiu2025multimodal}, which reuse or skip specific steps/blocks based on sampling-induced feature variations, and training-based methods \cite{ma2024learning}, where models learn to bypass non-critical modules. These are widely adopted in DiT architectures\cite{liu2025speca,zheng2025compute,zou2024accelerating} due to their consistent data dimensionality during sampling. U-Net-based methods like DeepCache \cite{ma2024deepcache} and Faster Diffusion \cite{li2023faster} achieve low-loss computation skipping via feature reuse. Adapting cache to DiTs is challenging, but Fora \cite{selvaraju2024fora} reuses attention or MLP layer outputs across denoising steps, and $\Delta$-DiT \cite{chen2024delta} targets specific blocks. 
EOC \cite{qiu2025acceleratingEOC} and GOC \cite{qiu2025acceleratingGOC} proposes method that determines when caching steps require optimization and leverages gradients in the caching process to reduce errors in future caching.


\section{Method}

In this section, we first introduce the preliminaries of controllable generative models, and then describe our acceleration strategy \textbf{H}ybrid-\textbf{G}rained \textbf{C}ache (\textbf{HGC}) in detail.


\subsection{Preliminaries}
Controllable generative models comprise two components: the control module and the generative module, where the control module processes conditional information and feeds it to the generative module for targeted content generation.

\textbf{Control Module} acts as the conditional interface, directing the generation process by processing external guidance signals, such as edge maps or segmentation masks. Given a conditional image ($\bm{I}^{\mathrm{C}}$), noise ($\bm{z}$), and prompt text ($\bm{P}$), it uses $T^{\mathrm{C}}$ calculation steps ($\bm{S}^{\mathrm{C}} = \{\bm{s}_{i}^{\mathrm{C}}\}_{i=1}^{T^{\mathrm{C}}}$) to encode these content, where each step contains an encoder and mid-block calculation layer ($\bm{s}_{i}^{\mathrm{C}} = [\bm{e}_{i}^{\mathrm{C}}, \bm{m}_{i}^{\mathrm{C}}]$). Thus, the control output ($\bm{O}_{i}^{\mathrm{C}}$) of each step can be formulated as:
\begin{equation}
\begin{aligned}
\bm{O}_{i}^{\mathrm{C}} &= [\bm{o}_{\bm{e},i}^{\mathrm{C}}, \bm{o}_{\bm{m},i}^{\mathrm{C}}]\\
         &= [\bm{e}_{i}^{\mathrm{C}}(\bm{I}^{\mathrm{C}}+ \bm{z}; \bm{P}), 
        \bm{m}_{i}^{\mathrm{C}}(\bm{e}_{i}^{\mathrm{C}}(\bm{I}^{\mathrm{C}}+ \bm{z}; \bm{P}))].
\end{aligned}
\end{equation}
Finally, the outputs are used to guide the generation process. 

\textbf{Generative Module}, typically a pre-trained model, handles the core synthesis process. It consists of $T^{\mathrm{G}}$ generation steps ($\bm{S}^{\mathrm{G}} = \{\bm{s}_{i}^{\mathrm{G}}\}_{i=1}^{T^{\mathrm{G}}}$), and each generation step contains encoder block $\bm{e}^{\mathrm{G}}$, mid-block $\bm{m}^{\mathrm{G}}$, and decoder block $\bm{d}^{\mathrm{G}}$. Therefore, the generation process can be defined as:
\begin{equation}
\bm{I}_{T^{\mathrm{G}}}^{\mathrm{G}} = p(\bm{I}_{T^{\mathrm{G}}}^{\mathrm{G}}|\bm{I}_{T^{\mathrm{G}}-1}^{\mathrm{G}})= \bm{s}_{T^{\mathrm{G}}}^{\mathrm{G}}(\bm{s}_{T^{\mathrm{G}}-1}^{\mathrm{G}}(\bm{z}; \bm{P})),
\end{equation}
where $\bm{s}^{\mathrm{G}}(\bm{z}; \bm{P}) = \bm{d}^{\mathrm{G}}([\bm{e}^{\mathrm{G}};\bm{m}^{\mathrm{G}}(\bm{e}^{\mathrm{G}}(\bm{z}; \bm{P}))])$, $p$ is a prediction function and $[;]$ denotes the concatenation of features along the channel dimension. As for controllable generation, the control information generated 
by the control module is used in each step of the calculation process. Specifically, $\bm{e}_{i}^{\mathrm{C}}$ and $\bm{m}_{i}^{\mathrm{C}}$ in $\bm{o}_{i}^{\mathrm{C}}$ are loaded into the existing $\bm{s}_i^{\mathrm{G}}$ calculation:
\begin{equation}
\bm{s}_{i}^{\mathrm{G}}(\bm{z}; \bm{P}) = \bm{d}_i^{\mathrm{G}}([ \bm{e}^{\mathrm{G}}+ \bm{e}_i^{\mathrm{C}};
(\bm{m}_i^{\mathrm{G}} + \bm{m}_i^{\mathrm{C}})(\bm{e}_i^{\mathrm{G}}(\bm{z}; \bm{P}))]).
\end{equation}

Our \textbf{H}ybrid-\textbf{g}rained \textbf{C}aches (\textbf{HGC}) is built on the controllable generative model. The pipeline is shown in Figure~\ref{fig:pipeline}. The Block-Level Cache focuses on the former steps of the control module, where the features stabilize and become less sensitive to changes. This cache strategy eliminates redundant computations, improving the overall efficiency of the model. In the generative module, the Block-Level Cache improves computational efficiency by controlling the frequency of updates to the model's blocks. The process is divided into two stages, with two parameters $\lambda_{\mathrm{intra}}$ and $\lambda_{\mathrm{inter}}$ governing the cache density at different levels. On the other hand, the Prompt-Level Cache is introduced to address redundancy in prompt signal processing. By reusing cross-attention features, the model reduces computational costs while preserving the fidelity of prompt-specific information throughout the generation process.

\subsection{Block-Level (Coarse-grained) Cache}

\noindent\textbf{Control Module} primarily focuses on the former half of the steps, and the calculation of the latter half has little effect on the overall generated results. Therefore, we directly discard computations in the latter steps and focus solely on the characteristics of features in the former steps. 

Denoting the output of all steps of the control module as $\mathcal{O}^{\mathrm{C}} = \{\bm{o}_{i}^{\mathrm{C}}\}_{i=1}^{T^{\mathrm{C}}}$, we calculate the similarity between the outputs before $\lfloor T^{\mathrm{C}}/2 \rfloor$ steps by cosine similarity:
\begin{equation}
\begin{aligned}
a_{i,j}^{\mathrm{C}} &= \mathrm{Cosine}(\bm{o}_{i}^{\mathrm{C}}, \bm{o}_{j}^{\mathrm{C}}), \\
&= \frac{1}{2} ({\frac{\langle \bm{o}_{\bm{e},i}^{\mathrm{C}}, \bm{o}_{\bm{e},j}^{\mathrm{C}} \rangle}{\|\bm{o}_{\bm{e},i}^{\mathrm{C}}\| \|\bm{o}_{\bm{e},i}^{\mathrm{C}}\|}} + {\frac{\langle \bm{o}_{\bm{m},i}^{\mathrm{C}}, \bm{o}_{\bm{m},j}^{\mathrm{C}} \rangle}{\|\bm{o}_{\bm{m},i}^{\mathrm{C}}\|   \|\bm{o}_{\bm{m},i}^{\mathrm{C}}\|}}),
\end{aligned}
\end{equation}
where $i=1,\cdots, \lfloor T^{\mathrm{C}}/2 \rfloor$ and $j=i + 1, \cdots, \lfloor T^{\mathrm{C}}/2 \rfloor$. $\langle \cdot, \cdot \rangle$ is the inner product between two vectors.

Then we find the most influential step $\tau^{\mathrm{C}}$ based on the similarity score between each step and all the subsequent steps. Specifically, denoting the similarity score set of the $i$-th step and subsequent steps as:
\begin{equation}
\bm{A}_i^C = a_{i,i+1}^{\mathrm{C}}, a_{i,i+2}^{\mathrm{C}}, \cdots, a_{i,\lfloor T^{\mathrm{C}}/2 \rfloor}^{\mathrm{C}}.
\end{equation}
Then, we determine whether all similarity scores in this step are greater than the threshold $\theta$: $a_i^C > \theta$, where $\forall a_i^C \in \bm{A}_i^C$. If the step satisfies all the conditions, we define this $i$-th step as the cached $\tau^{\mathrm{C}}$ step and use it in the control module. In other words, we cache the control output at step $ \tau^{\mathrm{C}} $ and replace the control outputs for all subsequent steps with the cached output $ \bm{o}_{\tau^{\mathrm{C}}}^{\mathrm{C}}$:

\begin{equation}
\bm{o}_j^{\mathrm{C}} \leftarrow \bm{o}_{\tau^{\mathrm{C}}}^{\mathrm{C}},  j=\tau^{\mathrm{C}} + 1,\cdots, \lfloor T^{\mathrm{C}}/2 \rfloor.
\end{equation}
If the $i$-th step does not satisfy the above conditions, we continue to judge the next $(i+1)$-th step until the $\lfloor T^{\mathrm{C}}/2 \rfloor$ step. This strategy helps us optimize the generation process by preventing redundant calculations in the control module.

\noindent\textbf{Generative Module} typically uses an interval-based cache strategy \cite{ma2024deepcache}. It updates the cache every $N$ time steps:
\begin{equation}
     \{\bm{s}_{i+1}^{\mathrm{G}}, \cdots, \bm{s}_{i+N-1}^{\mathrm{G}}\} \leftarrow \bm{s}_i^{\mathrm{G}},  i = 1, 1 + N, 1 + 2N, \cdots. 
\end{equation}
However, this full-scale brute-force caching will be unstable during the detailed calculation process. Therefore, we divide the generative module into the former and the latter parts, just like similar operations in the control module, and design two balance parameters ($\lambda_{\mathrm{intra}}$ and $\lambda_{\mathrm{inter}}$) to control the caching strength of different steps. Specifically, $\lambda_{\mathrm{intra}}$ is used to control the relative cache density between different blocks throughout the former steps. Thus, the cache interval for the mid-blocks $ \bm{m}^{\mathrm{G}} $ and decoder blocks $ \bm{d}^{\mathrm{G}} $ in the generative module is defined as:
\begin{equation}
    N_{\mathrm{intra}} = N \cdot \lambda_{\mathrm{intra}},
\end{equation}
where $ \lambda_{\mathrm{intra}} \in (0,1] $ controls the compression ratio. As $ \lambda_{\mathrm{intra}} $ increases, the frequency of updates for the control module output into the generative module decreases. Conversely, a smaller $ \lambda_{\mathrm{intra}} $ results in more frequent updates. Then, the cache used in the former steps can be formulated as two parts, one of the full cache of the encoder blocks:  
\begin{equation}
\{\bm{e}_{i+1}^{\mathrm{G}}, \cdots, \bm{e}_{i+N-1}^{\mathrm{G}}\} \leftarrow \bm{e}_i^{\mathrm{G}},  i = 1, 1 + N, 1 + 2N, \cdots. 
\end{equation}
And the partial cache of the mid-blocks  and decoder blocks:
\begin{equation}
\begin{aligned}
     \{\bm{m}_{i+1}^{\mathrm{G}}, \cdots, \bm{m}_{i+N_\mathrm{intra}-1}^{\mathrm{G}}\} &\leftarrow \bm{m}_i^{\mathrm{G}}, \\
     \{\bm{d}_{i+1}^{\mathrm{G}}, \cdots, \bm{d}_{i+N_\mathrm{intra}-1}^{\mathrm{G}}\} &\leftarrow \bm{d}_i^{\mathrm{G}}, 
\end{aligned}
\end{equation}
where $i = 1, 1 + {N_\mathrm{intra}}, 1 + 2{N_\mathrm{intra}}, \cdots$.

As for the latter steps, we introduce a scalable cache interval via a hyperparameter $\lambda_{\mathrm{inter}}$ as the generative module's high computational demands for refining fine-grained details limit its caching capacity. Thus, the cache operation used in the latter steps can be rewritten as:
\begin{equation}
     \{\bm{s}_{i+1}^{\mathrm{G}}, \cdots, \bm{s}_{i+N_\mathrm{inter}-1}^{\mathrm{G}}\} \leftarrow \bm{s}_i^{\mathrm{G}}, 
\end{equation}
where $N_{\mathrm{inter}} = N \cdot \lambda_{\mathrm{inter}}$, and $i = \lfloor T^{\mathrm{C}}/2 \rfloor, \lfloor T^{\mathrm{C}}/2 \rfloor + {N_\mathrm{inter}}, \lfloor T^{\mathrm{C}}/2 \rfloor + 2{N_\mathrm{inter}}, \cdots$. Similar to the $\lambda_\mathrm{intra}$, $\lambda_{\mathrm{inter}} \in (0,1]$ adjusts the cache interval ratio based on the module's computational requirements.



\begin{table*}[t]
\centering
\caption{Main results of three conditions on ControlNet, ControlNet++, where bold denotes the best performance, and underline indicates the second-best.}
\begin{tabular}{l|c|c|cc|ccc|cc|ccc}
\toprule
 & \multirow{2}{*}{\textbf{Dataset}} & \multirow{2}{*}{\textbf{Method}} & \multicolumn{5}{c|}{\textbf{ControlNet}} & \multicolumn{5}{c}{\textbf{ControlNet++}} \\ 
 &  &  & \textbf{FID↓} & \textbf{CS↑} & \textbf{MACs↓} & $\mathcal{L.}$\textbf{↓} & $\mathcal{S.}$ & \textbf{FID↓} & \textbf{CS↑} & \textbf{MACs↓} & $\mathcal{L.}$\textbf{↓} & $\mathcal{S.}$ \\ \midrule
\multirow{8}{*}{\rotatebox[origin=c]{90}{Segmentation}} & \multirow{4}{*}{ADE20k} & NoCache & \underline{38.13} & \textbf{31.10} & 18.22T & 5.24s & 1.00× & 30.51 & \textbf{31.96} & 18.22T & 5.24s & 1.00× \\
 &  & DeepCache & 38.74 & \underline{30.98} & \underline{9.30T} & \underline{3.03s} & \underline{1.73×} & 29.91 & \underline{31.84} & \underline{9.30T} & \underline{3.03s} & \underline{1.73×} \\
 &  & T-GATE & \textbf{37.92} & 30.79 & 13.51T & 4.27s & 1.23× & \textbf{28.12} & 31.77 & 13.51T & 4.27s & 1.23× \\
 &  & HGC & 39.44 & 30.26 & \textbf{6.70T} & \textbf{2.60s} & \textbf{2.02×} & \underline{28.93} & 31.57 & \textbf{6.70T} & \textbf{2.60s} & \textbf{2.02×} \\ \cmidrule(lr){2-13}
 & \multirow{4}{*}{COCOStuff} & NoCache & \underline{22.50} & \textbf{31.85} & 18.22T & 5.24s & 1.00× & \underline{19.99} & \textbf{32.37} & 18.22T & 5.24s & 1.00× \\
 &  & DeepCache & 22.98 & \underline{31.80} & \underline{9.30T} & \underline{3.03s} & \underline{1.73×} & 20.46 & \underline{32.36} & \underline{9.30T} & \underline{3.03s} & \underline{1.73×} \\
 &  & T-GATE & \textbf{22.15} & 31.40 & 13.51T & 4.27s & 1.23× & \textbf{19.03} & 31.99 & 13.51T & 4.27s & 1.23× \\
 &  & HGC & 22.93 & 30.82 & \textbf{6.70T} & \textbf{2.60s} & \textbf{2.02×} & 20.29 & 31.52 & \textbf{6.70T} & \textbf{2.60s} & \textbf{2.02×} \\ \midrule

\multirow{4}{*}{\rotatebox[origin=c]{90}{Edge}} & \multirow{4}{*}{Multi-20M} & NoCache & 21.22 & \textbf{32.13} & 18.22T & 5.24s & 1.00× & 20.14 & \underline{31.75} & 18.22T & 5.24s & 1.00× \\
 &  & DeepCache & \underline{20.69} & \underline{32.04} & \underline{9.30T} & \underline{3.03s} & \underline{1.73×} & 19.38 & \textbf{32.04} & \underline{9.30T} & \underline{3.03s} & \underline{1.73×} \\
 &  & T-GATE & \textbf{20.43} & 31.44 & 13.51T & 4.27s & 1.23× & \textbf{19.11} & 31.06 & 13.51T & 4.27s & 1.23× \\
 &  & HGC & 20.81 & 30.97 & \textbf{6.70T} & \textbf{2.60s} & \textbf{2.02×} & \underline{19.15} & 30.63 & \textbf{6.70T} & \textbf{2.60s} & \textbf{2.02×} \\ \midrule

\multirow{4}{*}{\rotatebox[origin=c]{90}{Depth}} & \multirow{4}{*}{Multi-20M} & NoCache & \underline{19.84} & \textbf{32.32} & 18.22T & 5.24s & 1.00× & \underline{15.94} & \textbf{32.33} & 18.22T & 5.24s & 1.00× \\
 &  & DeepCache & 20.58 & \underline{32.08} & \underline{9.30T} & \underline{3.03s} & \underline{1.73×} & 16.23 & \underline{32.29} & \underline{9.30T} & \underline{3.03s} & \underline{1.73×} \\
 &  & T-GATE & \textbf{19.54} & 31.22 & 13.51T & 4.27s & 1.23× & \textbf{14.39} & 31.51 & 13.51T & 4.27s & 1.23× \\
 &  & HGC & 21.87 & 30.55 & \textbf{6.70T} & \textbf{2.60s} & \textbf{2.02×} & 17.26 & 30.90 & \textbf{6.70T} & \textbf{2.60s} & \textbf{2.02×} \\ \bottomrule
\end{tabular}
\label{tab:main_result}
\end{table*}

\subsection{Prompt-Level (Fine-grained) Cache}
\noindent\textbf{Control Module} consists of the encoder block and mid block. These blocks have been proven to change minimally during the calculation \cite{li2023faster}. Thus, we opt to cache the cross-attention features at the first computation step of the control module. Denoting the cross-attention features of the first computation step in $\bm{s}_1^{\mathrm{C}}$ as $\bm{f}_1^{\mathrm{C}}$, we reuse the $\bm{f}_1^{\mathrm{C}}$ in the subsequent steps:
\begin{equation}
    \bm{f}_i^{\mathrm{C}} \leftarrow \bm{f}_{1}^{\mathrm{C}},  i = 1, \cdots, T^\mathrm{G}.
\end{equation}

\noindent\textbf{Generative Module} uses the cross-attention guiding the model to generate images aligned with the textual input. However, as described in T-GATE \cite{zhang2024cross}: the outputs of cross-attention tend to converge after a few inference steps, leading to redundancy in its computational use during the latter steps of image generation. Thus, we introduce a gate step $\tau^{\mathrm{G}}=\lfloor T^{\mathrm{G}}/2 \rfloor$. Here, we split the data fed into the model into two parts, one for prompt calculation and the other for non-prompt calculation. We fuse the results of these two parts in the $\tau^{\mathrm{G}}$ step. Denoting the results of the cross-attention combined with the prompt and without the prompt as $\bm{f}_{\mathrm{p},\tau^{\mathrm{G}}}^{\tau^{\mathrm{G}}}$ and $\bm{f}_{\mathrm{np},\tau^{\mathrm{G}}}^{\tau^{\mathrm{G}}}$, respectively, the fusion operation can be formulated as:
\begin{equation}
    \bm{f}_\mathrm{fuse}^{\mathrm{G}} = (\bm{f}_{\mathrm{p},\tau^{\mathrm{G}}}^{\tau^{\mathrm{G}}} + \bm{f}_{\mathrm{np},\tau^{\mathrm{G}}}^{\tau^{\mathrm{G}}})/2.
\end{equation}
For subsequent steps $i = \tau^{\mathrm{G}}, \tau^{\mathrm{G}}+1, \cdots, T^{\mathrm{G}}$, we reuse $\bm{f}_\mathrm{fuse}^{\mathrm{G}}$ to reduce the calculation cost:
\begin{equation}
    \bm{f}_i^{\mathrm{G}} \leftarrow \bm{f}_\mathrm{fuse}^{\mathrm{G}}.
\end{equation}

To further enhance efficiency, we halve the batch size in Control and Generative modules during the fusion phase, significantly reducing memory overhead without compromising generation quality. Specifically, During inference with batch size $n$, the actual processing requires $2n$ computations ($n$ for $\bm{f}_{\mathrm{p},\tau^{\mathrm{G}}}^{\tau^{\mathrm{G}}}$ and $n$ for $\bm{f}_{\mathrm{np},\tau^{\mathrm{G}}}^{\tau^{\mathrm{G}}}$). At step $\tau^{\mathrm{G}}$, we cache both attention maps and subsequently replace the dual computations with a fused version $\bm{f}_\mathrm{fuse}^{\mathrm{G}}$, this fusion reduces the computational batch size from $2n$ back to $n$, halving the batch size. Notably, for subsequent steps $i = \tau^{\mathrm{G}}, \tau^{\mathrm{G}}+1, \cdots, T^{\mathrm{G}}$, although the cache interval decreases to accommodate fine-grained detail refinement, the actual computational overhead remains minimal due to the batch size halving operation.

\section{Experiment}
In this section, we conduct experiments to evaluate the performance of our proposed hybrid-grained cache (\textbf{HGC}). We intended to address the following research questions (\textbf{RQ}):

\noindent\textbf{RQ1:} Is Hybrid-grained Caches (HGC) effective?

\noindent\textbf{RQ2:} What hyperparameter should be selected?

\noindent\textbf{RQ3:} Can HGC be applied to other models?

\subsection{Experiment Settings}

\noindent\textbf{Models and Datasets.}
We utilize two baseline models in our experiments: ControlNet \cite{zhang2023adding} and ControlNet++ \cite{li2024controlnet++}. These models have similar architectures, including control modules and generation modules. We tested our method on several datasets: ADE20K \cite{zhou2017scene} and COCOStuff \cite{caesar2018coco} for generation with segmentation mask conditions, and the MultiGen-20M dataset from UniControl \cite{qin2023unicontrol}, a subset of LAION-Aesthetics \cite{schuhmann2022laion}, for generation of canny edge map and depth map conditions. For datasets lacking text captions, such as ADE20K, we use MiniGPT4 \cite{zhu2023minigpt} to generate one sentence image caption with the prompt: ``Please briefly describe this image in one sentence.'' Then, we use this caption as the prompt to evaluate our method.

\noindent\textbf{Cache Settings.}
In our method, we set the number of steps in the control module \( T^{\mathrm{C}} \) and the generative module \( T^{\mathrm{G}} \) to 20. For Block-level Cache, we set the similarity threshold \( \theta \) to 0.9, which results in a cached step \( \tau^{\mathrm{C}} = 6 \). In the generative module, we use a base cache interval \( N = 5 \), with \( \lambda_{\mathrm{intra}} = 0.4 \) and \( \lambda_{\mathrm{inter}} = 0.6 \) to control the caching behavior. For Prompt-level Cache, we get the gate step \( \tau^{\mathrm{G}} = 10 \), which determines when the model switches to caching the fused attention maps.
Additionally, for DeepCache \cite{ma2024deepcache}, we set its cache interval to 5, and for T-GATE \cite{zhang2024cross}, we set the gate step to 5. The complete set of hyperparameters and implementation details is available in the source code.


\noindent\textbf{Evaluation.}
The images are generated using DDIM \cite{song2020denoising} with a predefined 20 inference steps with guiding scale (7.5) and resized to 512 × 512 resolution. 
We employed metrics such as Fr'{e}chet Inception Distance (\textbf{FID}) and CLIP Score (\textbf{CS}) to measure the quality of generated images. To evaluate the efficiency, we use Calflops to count Multiple-Accumulate Operations (\textbf{MACs}). Furthermore, we measure the end-to-end latency ($\mathcal{L.}$) and Speedup ($\mathcal{S.}$) for processing a batch of 4 samples on a system powered by one NVIDIA RTX 3090 GPU.

\subsection{Main Results (RQ1)}
We employed two generative models with three distinct conditions (segmentation, edge, and depth maps) for our experiments. Results are presented in Table \ref{tab:main_result}. We can find that our HGC helps the model achieve a speed improvement while maintaining comparable generation quality. Specifically,
compared to the 1.73x acceleration of DeepCache, HGC achieves a 2.02x acceleration (16.8\% $\uparrow$). As for quality analysis, for the metric FID, the gap between HGC and DeepCache is controlled within 2\% for most tasks, with only the depth task showing a gap of around 6\%. 

\begin{table}[t]
\centering
\caption{Comparison of different acceleration methods.}
\begin{tabular}{c|ccc}
\toprule
\textbf{Method} & \textbf{MACs↓} & \textbf{FID↓} & \textbf{CS↑} \\ \midrule
NoCache & 18.22T & \textbf{19.99} & \textbf{32.37} \\
DeepCache & 8.03T & 24.84 & 31.68 \\
T-GATE & 10.23T & 17.32 & 30.88 \\
HGC & \textbf{6.70T} & 20.29 & 31.52 \\ \bottomrule
\end{tabular}
\label{tab:method_comparison}
\end{table}
Meanwhile, we aligned the acceleration ratios of T-GATE, DeepCache, and HGC to compare the quality of images generated using the COCO-Stuff-Seg dataset with ControlNet. We ensured that the computational requirements (MACs) for T-GATE, DeepCache, and HGC were nearly identical. As shown in Table~\ref{tab:method_comparison}, HGC outperforms other methods in generation quality under similar computational budgets. Although T-GATE achieves the best FID score, its CLIP Score is lower, indicating reduced alignment between generated images and their prompts. Notably, we set T-GATE's gate step to 3, exceeding the recommended range in the original T-GATE paper, suggesting HGC offers greater acceleration potential than both T-GATE and DeepCache.

\begin{table}[t]
\caption{Comparison of different threshold $\theta$.}
\centering
\begin{tabular}{c|c|cc|c}
\toprule
$\theta$ & $\bm{\tau}^{\mathrm{C}}$ & \textbf{FID↓} & \textbf{CS↑} & \textbf{MACs↓} \\ \midrule
0 & 1 & 30.92 & 31.26 & \textbf{5.53T} \\ 
0.9 & 6 & 28.93 & \textbf{31.57} &  6.70T  \\ 
1.0 & 10 & \textbf{28.72} & 31.56 & 7.63T \\ \bottomrule
\end{tabular}
\label{tab:choices_theta}
\end{table}

\begin{table}[t]
\caption{Comparison of different parameter $\lambda_{\mathrm{intra}}$.}
\centering
\begin{tabular}{c|ccc}
\toprule
$\lambda_{\mathrm{intra}}$ & \textbf{FID↓} & \textbf{CS↑} & \textbf{MACs↓} \\ \midrule
0 & \textbf{28.86} & \textbf{31.68} & 8.418T \\
0.4 & 28.93 & 31.57 & 6.70T \\ 
1.0 & 29.04 & 31.49 & \textbf{5.826T} \\ \bottomrule
\end{tabular}
\label{tab:choice_intra}
\end{table}

\begin{table}[t]
\caption{Comparison of different parameter $\lambda_{\mathrm{inter}}$.}
\centering
\begin{tabular}{c|ccc}
\toprule
$\lambda_{\mathrm{inter}}$ & \textbf{FID↓} & \textbf{CS↑} & \textbf{MACs↓} \\ \midrule
0 & \textbf{28.44} & \textbf{31.64} & 8.61T \\
0.6 & 28.93 & 31.57 & 6.70T \\ 
1.0 & 30.49 & 31.40 & \textbf{6.06T} \\ \bottomrule
\end{tabular}
\label{tab:choice_inter}
\end{table}

\begin{figure*}[t]
  \centering
   \includegraphics[width=0.9\linewidth]{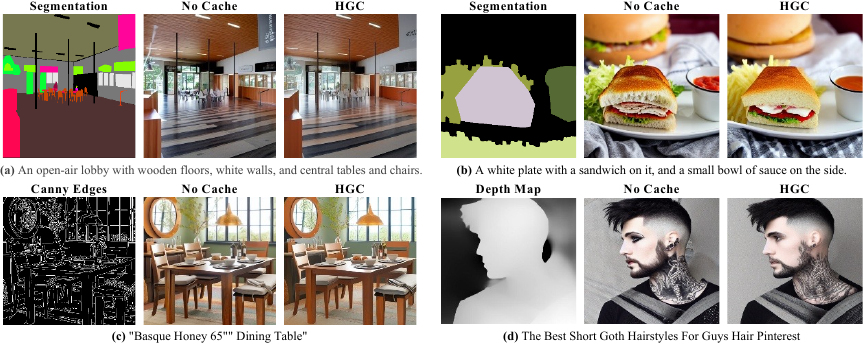}
   \caption{The visualization of the generation with or without HGC: (a) and (b) generation with segmentation condition. (c) generation with edge map. (d) generation with the depth map.
   \vspace{-0.15cm}
   }
   \label{fig:Quantitative_seg}
\end{figure*}

\begin{table}[t]
\caption{Comparison of cache strategy performance across different cache components, where GM denotes the generative module and CM refers to the control module.}
\centering
\begin{tabular}{l|ccc}
\toprule
\textbf{Method} & \textbf{MACs↓} & \textbf{FID↓} & \textbf{CS↑} \\ 
\midrule
NoCache & 18.22T & 19.99 & 32.37 \\
GM Block & 12.09T & 19.75 & \textbf{32.40} \\
CM Block & 14.95T & 20.86 & 32.22 \\ 
GM Prompt & 13.51T & \textbf{19.14} & 31.99 \\
CM Prompt & 15.77T & 20.27 & 32.19 \\
HGC & \textbf{6.70T} & 19.15 & 31.52 \\
\bottomrule
\end{tabular}
\label{tab:caching_strategies}
\end{table}

\begin{table}[t]
\caption{Comparison of video generation performance between baseline and HGC approaches.}
\centering
\begin{tabular}{l|ccc}
\toprule
\textbf{Method} & \textbf{MACs↓} & \textbf{FID↓} \\ 
\midrule
NoCache & 48.29T & 11.03  \\
\textbf{HGC} & 28.88T & 11.70  \\
\bottomrule
\end{tabular}
\label{tab:video_results}
\end{table}


\subsection{Ablation Studies (RQ2)}
In this section, we conduct experiments on ADE20k with segmentation mask and analyze the impact of various hyperparameters in our method. Specifically, we vary the target parameters while keeping all other parameters constant.



\noindent\textbf{Selection of $\theta$.}
As shown in Table \ref{tab:choices_theta}, we evaluate threshold values $\theta$ $\in \{0, 0.9, 1.0\}$, which yield corresponding caching steps $\tau^{\mathrm{C}}$ $= \{0, 0.4, 1.0\}$. 
When $\theta$ is set too low, the quality of the generated images significantly decreases, as evidenced by the FID increasing from 28.93 to 30.92, a increase of 6.88\%. This suggests that when threshold is too small, the model reaches the caching step too early, leading to less precise feature processing and a noticeable decline in image quality. On the other hand, setting $\theta$ too high results in computational redundancy. For example, increasing $\theta$ from 0.9 to 1.0 increases the computational cost (MACs) by 1T, but the image quality remains almost unchanged, with FID only improving slightly from 28.93 to 28.72. This indicates that excessively large thresholds lead to unnecessary computations, with diminishing returns in image quality.

\noindent\textbf{Selection of $\lambda_{\mathrm{intra}}$.}
As shown in Table \ref{tab:choice_intra}, we evaluate three distinct values of \( \lambda_{\mathrm{intra}} \) $\in \{0, 0.4, 1.0\}$. Quantitative analysis reveals a consistent performance degradation trend: the FID score increases from 28.86 to 29.04 while the CLIP Score decreases from 31.68 to 31.49 as \( \lambda_{\mathrm{intra}} \) varies from 0 to 1.0. This inverse correlation between \( \lambda_{\mathrm{intra}} \) and generation quality suggests that excessive dependency on cached features (higher \( \lambda_{\mathrm{intra}} \)) compromises the model's ability to maintain optimal image fidelity.

\noindent\textbf{Selection of $\lambda_{\mathrm{inter}}$.}
As shown in Table \ref{tab:choice_inter}, we evaluate three values 0, 0.6 and 1.0 for the ratio \( \lambda_{\mathrm{inter}} \). When \( \lambda_{\mathrm{inter}} \) increases, the image quality decreases. Specifically, when \( \lambda_{\mathrm{inter}} \) increases from 0 to 0.6, the change in quality is minimal, with the FID increasing slightly from 28.44 to 28.93 and the CLIP Score decreasing from 31.64 to 31.57. However, when \( \lambda_{\mathrm{inter}} \) increases from 0.6 to 1.0, there is a significant drop in image quality, with FID increasing to 30.49 and CLIP Score decreasing further to 31.40. This indicates that higher values of \( \lambda_{\mathrm{inter}} \) lead to a more severe decline in generation quality, while speed only improves slightly.

\noindent\textbf{Different Caching Components.}
As shown in Table \ref{tab:caching_strategies}, all four independent components successfully reduced computational costs while maintaining generation quality comparable to the Baseline. From an internal comparison perspective, the acceleration achieved for the generative module at the same granularity level was more substantial than that for the control module, which aligns with the fact that the generative module accounts for the majority of computations in Controllable Generation. From a granularity perspective, coarse-grained approaches demonstrated greater acceleration potential compared to fine-grained methods.

\subsection{Exploration on Video Generation (RQ3)}
We extend the evaluation of our HGC method to video generation tasks by integrating it with the CTRL-Adapter framework \cite{lin2024ctrl}, using DAVIS 2017 dataset \cite{pont20172017}. As shown in Table \ref{tab:video_results}, our approach achieves a significant 40\% reduction in computational cost (from 48.29T to 28.88T MACs for 14-frame generation) while maintaining reasonable output quality, with the FID score increasing from 11.03 to 11.70. This acceleration demonstrates the effectiveness of our caching strategy for video generation tasks, where the trade-off between computational efficiency and visual fidelity remains within acceptable limits. Results suggest that our method can be successfully adapted to sequential generation tasks while preserving its core advantages in computational reduction.

\subsection{Visualization} 
Visualization results are shown in Figure \ref{fig:Quantitative_seg}, which includes: (a) ADE20K dataset with segmentation masks, (b) COCO-Stuff dataset with segmentation masks,
(c) MultiGen-20M dataset with Canny edges, and (d) MultiGen-20M dataset with depth maps. Our comparative analysis under identical control images and prompts reveals that NoCache and HGC methods generate images that faithfully align with the structural and thematic requirements of the input constraints. However, close inspection highlights HGC's subtle trade-off between efficiency and micro-detail fidelity: while it retains macro-structural integrity, it exhibits reduced precision in high-frequency details such as wood grain textures, shadow gradations around televisions, and fine textural patterns in clothing. This discrepancy stems from HGC's block-level cache mechanism—reusing intermediate features across denoising steps inherently smooths out transient details accumulated through iterative refinement. Despite this, HGC achieves comparable visual coherence to NoCache in all evaluated scenarios, successfully balancing computational efficiency with perceptually acceptable quality degradation.

\section{Conclusion}
In this work, we propose HGC, a dual-level cache framework that accelerates controllable generation through joint optimization of prompt-level and block-level cache mechanisms. Experiments demonstrate that HGC achieves nearly 2× speedup across diverse control tasks while preserving competitive output quality. However, limitations emerge in geometrically complex scenarios, where a 3–5\% CLIP Score degradation occurs due to insufficient adaptability of cached cross-attention maps during rapid scene transitions. These challenges highlight opportunities for future enhancements, particularly in developing variance-aware adaptive interval scheduling and attention-guided dynamic cache invalidation strategies. In our future work, we plan to explore more general accelerating techniques to reduce the performance loss.

\section{Acknowledgments}
This research is supported by the National Natural Science Foundation of China (No.U24B20180, No. 62576330, No.62472393) , National Natural Science Foundation of Anhui (No.2508085MF143) and the advanced computing resources provided by the Supercomputing Center of the USTC.

\bibliography{aaai2026}

\end{document}